\documentclass[sigconf]{acmart}

\AtBeginDocument{%
  }

\usepackage{microtype}
\usepackage{graphicx}
\usepackage{subcaption}
\usepackage{booktabs} 

\usepackage{hyperref}
\usepackage{enumitem}
\usepackage{balance}

\usepackage{multirow}
\usepackage{multicol}

\usepackage[T1]{fontenc}

\newcommand{\stdfont}{\fontsize{8pt}{8pt}\selectfont}

\copyrightyear{2022}
\acmYear{2022}
\setcopyright{acmlicensed}
\acmConference[CIKM '22]{Proceedings of the 31st ACM International Conference on Information and Knowledge Management}{October 17--21, 2022}{Atlanta, GA, USA}
\acmBooktitle{Proceedings of the 31st ACM International Conference on Information and Knowledge Management (CIKM '22), October 17--21, 2022, Atlanta, GA, USA}
\acmPrice{15.00}
\acmDOI{10.1145/3511808.3557681}
\acmISBN{978-1-4503-9236-5/22/10}

\begin{document}

\title[ReFine: Re-randomization before Fine-tuning for Cross-domain Few-shot Learning]{ReFine: Re-randomization before Fine-tuning\\for Cross-domain Few-shot Learning}

\author{Jaehoon Oh}
\authornote{The authors contributed equally to this research.}
\orcid{0000-0002-4298-1762}
\affiliation{
  \institution{KAIST DS}
  \city{Daejeon}
  \country{Republic of Korea}
  \postcode{34141}
}
\email{jhoon.oh@kaist.ac.kr}

\author{Sungnyun Kim}
\orcid{0000-0002-3251-1812}
\authornotemark[1]
\affiliation{%
  \institution{KAIST AI}
  \city{Seoul}
  \country{Republic of Korea}
  \postcode{02455}
}
\email{ksn4397@kaist.ac.kr}

\author{Namgyu Ho}
\orcid{0000-0002-2445-3026}
\authornotemark[1]
\affiliation{%
  \institution{KAIST AI}
  \city{Seoul}
  \country{Republic of Korea}
  \postcode{02455}
}
\email{itsnamgyu@kaist.ac.kr}

\author{Jin-Hwa Kim}
\orcid{0000-0002-0423-0415}
\affiliation{
  \institution{NAVER AI Lab}
  \city{Sungnam}
  \country{Republic of Korea}
  \postcode{13561}
}
\email{j1nhwa.kim@navercorp.com}

\author{Hwanjun Song}
\authornote{Corresponding authors}
\orcid{0000-0002-1105-0818}
\affiliation{
  \institution{NAVER AI Lab}
  \city{Sungnam}
  \country{Republic of Korea}
  \postcode{13561}
}
\email{hwanjun.song@navercorp.com}

\author{Se-Young Yun}
\orcid{0000-0001-6675-5113}
\authornotemark[2]
\affiliation{%
  \institution{KAIST AI}
  \city{Seoul}
  \country{Republic of Korea}
  \postcode{02455}
}
\email{yunseyoung@kaist.ac.kr}





\renewcommand{\shortauthors}{Jaehoon Oh et al.}

\begin{abstract}
Cross-domain few-shot learning (CD-FSL), where there are few target samples under extreme differences between source and target domains, has recently attracted huge attention.
Recent studies on CD-FSL generally focus on transfer learning based approaches, where a neural network is pre-trained on popular labeled source domain datasets and then transferred to target domain data.
Although the labeled datasets may provide suitable initial parameters for the target data, the domain difference between the source and target might hinder fine-tuning on the target domain.
This paper proposes a simple yet powerful method that re-randomizes the parameters fitted on the source domain before adapting to the target data.
The re-randomization resets source-specific parameters of the source pre-trained model and thus facilitates fine-tuning on the target domain, improving few-shot performance.
\end{abstract}





\maketitle

\section{Introduction}


Few-shot learning (FSL) has become an attractive field of deep learning research to tackle problems with a small number of training samples \citep{wang2020generalizing}.
In this setting, a model is typically pre-trained on a large source dataset comprised of \emph{base} classes from the source domain and then transferred into the target dataset comprised of few samples from unseen \emph{novel} classes.
Studies on FSL have typically assumed that the base and novel classes share the same domain, and these have followed two research directions: meta-learning \citep{maml, snell2017prototypical, raghu2019rapid, oh2021boil} and fine-tuning \citep{chen2018a, Dhillon2020A, tian2020rethinking}.

However, the source dataset and the target dataset come from considerably different domains in many real-world scenarios \citep{bscd_fsl, phoo2021selftraining}.
To tackle this problem, \emph{cross-domain few-shot learning} (CD-FSL) has recently gained significant attention, exemplified by the introduction of the BSCD-FSL benchmark dataset\,\citep{bscd_fsl}. This benchmark considers large-scale natural image datasets as source data and four different target datasets for evaluation, each with varying levels of similarity to the source data domain. It is shown that transfer learning approaches, where a pre-trained model on the source domain is fine-tuned on the target domain, overwhelm meta-learning approaches on BSCD-FSL \cite{bscd_fsl}.




\begin{figure}[!t]
\centering
\includegraphics[width=1.0\linewidth]{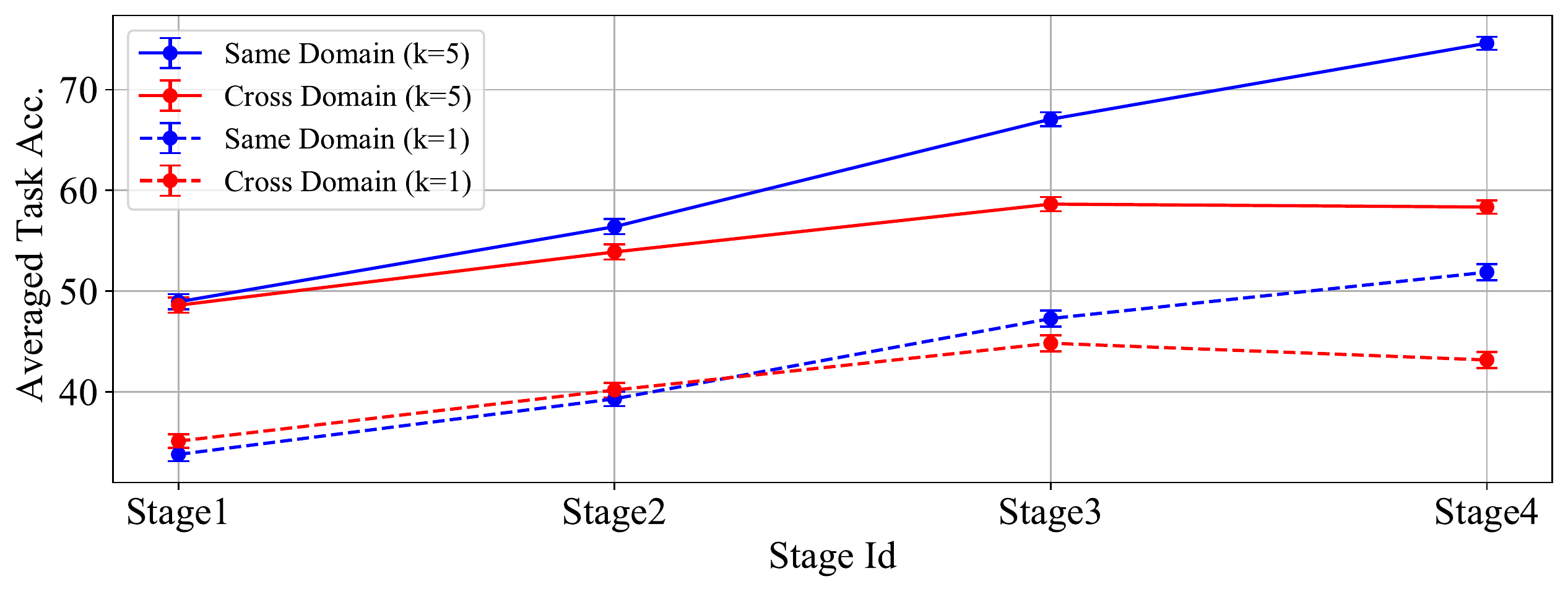}
\caption{FSL accuracy (5-way $k$-shot) using the intermediate representation from each stage in ResNet10 (refer to Figure \ref{fig:backbone} for the ResNet10 structure). An average pooling layer and an auxiliary classifier are attached at the end of each stage. After pre-training on miniImageNet (source domain), the model is transferred to the same domain (blue lines) or four different target domains in the BSCD-FSL benchmark (red lines), where only the attached classifier is fine-tuned. The cross-domain accuracy is averaged on the four domains.}
\vspace{-10pt}
\label{fig:motivation}
\end{figure}


In this regard, recent works have attempted to extract better representations during the pre-training phase by exploiting unlabeled data from the target domain \citep{phoo2021selftraining, islam2021dynamic, oh2022understanding} or reconstructing the images with an autoencoder to enhance the generalization of a model \citep{liang2021boosting}.
While these works focus on developing better pre-training methods, we suppose the fine-tuning phase is also a crucial research direction.
\citet{das2021confess} were aware of the importance of fine-tuning for CD-FSL, however, their framework using a mask generator is highly complicated to use.

In this paper, we present a {new} perspective to tackle the domain gap issue in CD-FSL: \emph{not all the pre-trained parameters from the source domain are desirable on the target domain}.
We posit that parameters in deeper layers of a pre-trained feature extractor may be detrimental for target domain adaptation, as they contain domain-specific information belonging to the source domain.
This is demonstrated in Figure \ref{fig:motivation}, where we use fixed image features from different stages of a pre-trained backbone and analyze the change in few-shot performance. We observe different trends for same-domain and cross-domain scenarios. While accuracy increases consistently with feature depth in the same-domain case (the blue lines), the accuracy decreases when using features from the last stage in the cross-domain case (the red lines).

Motivated by these findings, we propose a novel method, \textbf{ReFine} (\underline{Re}-randomization before \underline{Fine}-tuning), where we {re-randomize} the top layers of the feature extractor after supervised training on the source domain, before fine-tuning on the target domain.
%
This is effective for CD-FSL because it helps reduce the learning bias towards the source domain by simply re-randomizing the domain-specific layer. It can also be implemented by adding a few lines of code and can be easily combined with other recent CD-FSL methods. This {simplicity} and {flexibility} allows it to be easily adapted in practical uses for CD-FSL. 
Contrary to the prior works that have focused on improving universal representations during the pre-training phase\,\cite{phoo2021selftraining, oh2022understanding}, our method focuses on removing source-specific features obtained during pre-training to aid the fine-tuning.

Our contributions are summarized as follows:
\begin{itemize}[leftmargin=*]
    \item We propose a simple yet effective algorithm called ReFine, which re-randomizes the fitted parameters on the source domain and then fine-tunes the partially re-randomized model. This puts forward a new perspective for adapting to novel classes of the target domain in CD-FSL.
    \item We demonstrate improved performance for CD-FSL when our re-randomization technique is used, and provide an in-depth analysis on where and how to re-randomize.
\end{itemize}
\begin{figure}[!t]
  \centering
  \includegraphics[width=\linewidth]{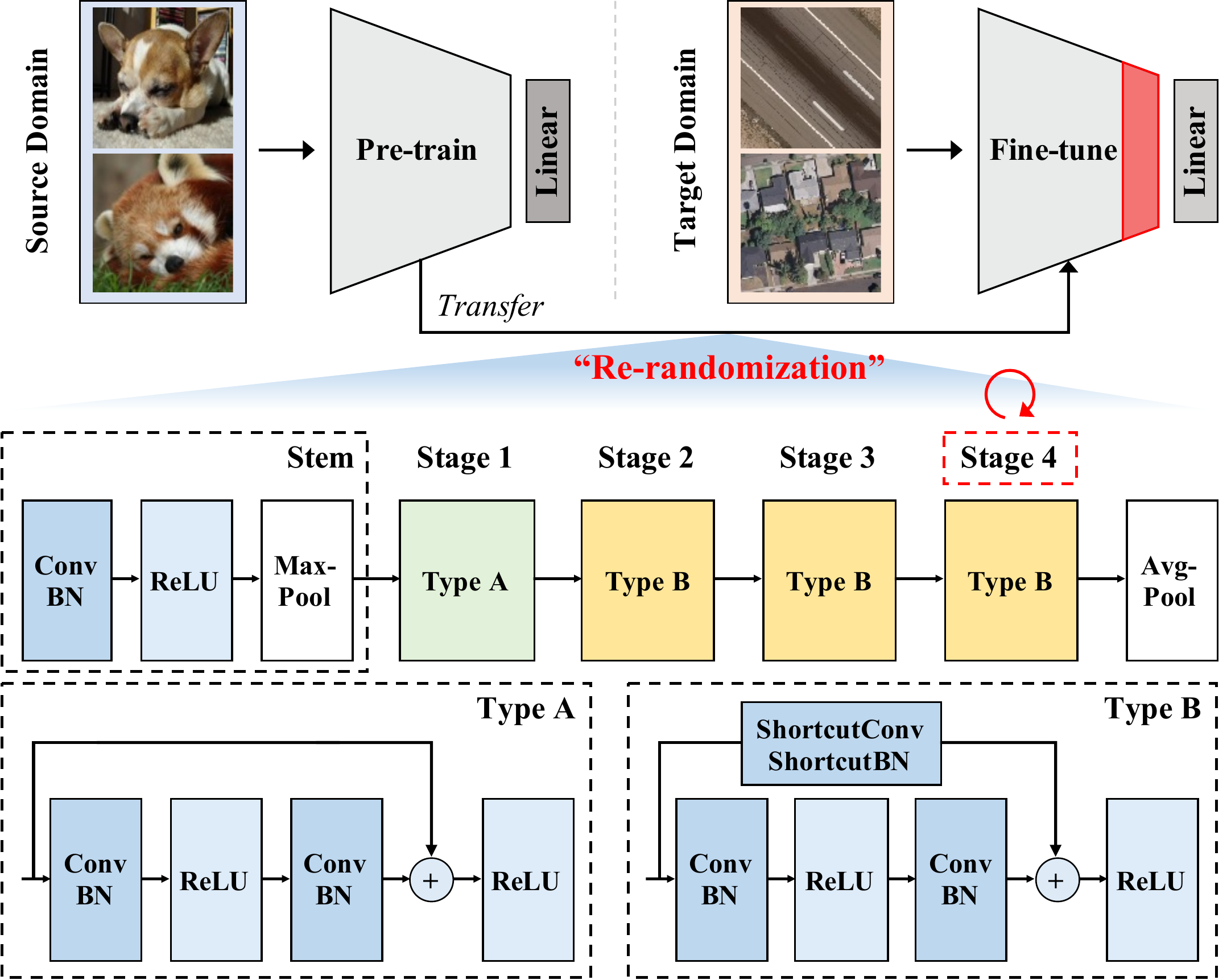}
  \caption{Overview of our proposed algorithm, ReFine, with the structure of ResNet10 backbone network.}
  \label{fig:backbone}
\end{figure}

\section{Related Works}\label{sec:related_works}

\textbf{Few-shot learning (FSL)} has been studied in two research directions, meta-learning and fine-tuning. Regarding the meta-learning approach, a meta-trained model is evaluated after fast adaptation on a few train sets. The meta-training procedure resembles the episodic evaluating procedure. Meta-learning approaches include learning good initialized parameters \cite{maml, pmaml, tian2020rethinking, oh2021boil}, a metric space \cite{matchingnet, snell2017prototypical, relationnet, chen2021multi}, and update rule or optimization \cite{metasgd, learntolearn, Flennerhag2020Meta-Learning}. Regarding the fine-tuning approach \cite{chen2018a, Dhillon2020A, tian2020rethinking}, a pre-trained model is typically evaluated after fine-tuning.
During the pre-training procedure, the model is trained in a mini-batch manner.

\medskip
\noindent\textbf{Cross-domain few-shot learning (CD-FSL)} addresses a problem when the source and target domains are extremely different, which is a more real-world scenario for FSL \cite{bscd_fsl, feature_transformation}.
Initially, \citet{feature_transformation} proposed feature-wise transformation (FWT) that learns scale-and-shift meta-parameters using pseudo-unseen target data during meta-training. 
However, it showed poor performance on the recently released BSCD-FSL benchmark \cite{bscd_fsl}, consisting of four target datasets collected from different domains.
%
In general, fine-tuning based approaches have been shown to outperform meta-learning based approaches such as FWT \cite{bscd_fsl}. Therefore, recent CD-FSL studies have proposed their algorithms under a pre-training and fine-tuning scheme.
These works have mainly focused on improving the pre-training phase, so that the pre-trained model is more suitable for adaptation to the target domain.

\medskip
\noindent\textbf{Re-randomization}\footnote{Although some literature use the term \textit{re-initialization}, we distinguish it from \textit{re-randomization} because \textit{re-initialization} reverts the values to the previously initialized ones. Refer to \cite{zhang2019all} for a formal definition. For a more concrete comparison, we have also dealt with \textit{re-initialization} in Section \ref{subsubsec:how_to_reinit}.} has been widely studied in the field of language tasks \cite{bert_finetune, investigate_transfer}, in particular related to BERT, which is one of the most popular fine-tuning based language models.
An interesting observation from \citet{bert_finetune} is that re-randomizing the topmost block in BERT increases the performance for downstream tasks by reducing the fine-tuning workload. 
%
%
Concurrently to this observation, \citet{investigate_transfer} examined the relations between the partial re-randomization of BERT and transferability of the layers.
Meanwhile, in a visual task, \citet{reinit_cnn} showed that placing more emphasis on the early layers of a convolutional neural network helps improve generalization.
%
There have been similar attempts in meta-learning based FSL, e.g., zeroing the context vector for adaptation in each new task \cite{cavia} and setting the classifier weight to have the same row vector (for any-shot problem) \cite{hidra}.
However, to the best of our knowledge, our work is the first to investigate the impact of re-randomization in fine-tuning based approaches for better CD-FSL.

\section{ReFine: Re-randomization before Fine-tuning}\label{sec:problem_statement}
The objective of fine-tuning based CD-FSL algorithms is to learn a backbone $f$ on the source data $D_B$ with base classes $C_B$, extracting meaningful representations on the target data $D_N$ with novel classes $C_N$, where $C_B \cap C_N = \emptyset$.
However, because there is no access to target data, the pre-trained model is biased towards the source domain, especially in the upper layers that pertain to classification of base classes.
To mitigate this, we re-randomize the upper layers of the pre-trained backbone $f$ to reset source-fitted parameters, depicted in Figure \ref{fig:backbone}.
Specifically, the weights of convolutional layers are re-randomized to uniform distributions \cite{he2015delving}.
The scaling and shifting parameters of batch normalization layers are reset to ones and zeros, respectively.
%

The reason why \emph{upper} layers of the backbone $f$ are re-randomized is that more domain-specific representations are extracted as the depth increases in convolutional neural networks \cite{yosinski2014transferable, raghu2019transfusion, maennel2020neural, reinit_cnn}. Re-randomization of upper layers helps the training loss to escape from local minima attributed to $D_B$ and allows bottom-level layers to be sufficiently updated, alleviating the gradient vanishing problem \cite{ro2019backbone, li2020rifle}.
This is in line with previous works which show that representation change is beneficial for CD-FSL \cite{feature_transformation, oh2021boil}.


Finally, fine-tuning and evaluation are performed with episodes, each representing distinct tasks, sampled from the labeled target data $D_N$. Each episode consists of a support set $D_s$, used to fine-tune the partially re-randomized pre-trained model, and a query set $D_q$, used to evaluate after the fine-tuning. To sample an episode $(D_s, D_q)$, $n$ classes are first selected from $C_N$, and subsequently, $k$ and $k_q$ samples are selected per class for support and query sets, respectively, where $n=5$ and $k \in \{1,5\}$ in general.
\section{Experiments}\label{sec:experiments}

We introduce the experimental setup in Section \ref{subsec:setup} and compare \textbf{ReFine} (ours) with two baselines in Section \ref{subsubsec:only_rerand_comp}: (1) \textbf{Linear} is a linear probing method to fine-tune only the classifier layer; (2) \textbf{Transfer} is a transfer learning method to fine-tune the entire network without using re-randomization\footnote{Many meta-learning based approaches such as MAML, ProtoNet, ProtoNet+FWT, and MetaOptNet have worse performance than Transfer, which is shown in \cite{bscd_fsl}.}.  We further investigate where and how to re-randomize in Section \ref{subsubsec:where_to_reinit} and Section \ref{subsubsec:how_to_reinit}, respectively.

\begin{table}[t]
\centering
\caption{5-way $k$-shot accuracy over 600 tasks on \{miniIN, tieredIN\}\,$\rightarrow$\,\{BSCD-FSL\}. For ReFine, topmost layers in the last stage are re-randomized (see Section \ref{subsubsec:where_to_reinit}). Mean and 95\% confidence interval are reported.}\label{tab:BSCD_comparison_all}
\vspace{-5pt}
\addtolength{\tabcolsep}{-4pt}
\resizebox{\linewidth}{!}{%
\begin{tabular}{ll|cccc}
    \toprule
     & & \multicolumn{4}{c}{Target dataset} \\
    & & $k=1$ & $k=5$ & $k=1$ & $k=5$ \\
    \cmidrule{3-6}
    Source dataset & Methods & \multicolumn{2}{c}{CropDisease} & \multicolumn{2}{c}{EuroSAT} \\
    \midrule
    \multirow{3}{*}{\shortstack{miniImageNet}} & Linear   &  65.73{\stdfont$\pm$.87} & 88.68{\stdfont$\pm$.53} & 54.35{\stdfont$\pm$.92} & 75.96{\stdfont$\pm$.67} \\
    & Transfer & 57.57{\stdfont$\pm$.92} & 88.04{\stdfont$\pm$.57} & 51.54{\stdfont$\pm$.86} & 79.33{\stdfont$\pm$.66} \\ 
    & ReFine & \textbf{68.93{\stdfont$\pm$.84}} & \textbf{90.75{\stdfont$\pm$.49}} & \textbf{64.14{\stdfont$\pm$.82}} & \textbf{82.36{\stdfont$\pm$.57}} \\
    \midrule
    \multirow{3}{*}{\shortstack{tieredImageNet}} & Linear & \textbf{70.88{\stdfont$\pm$.90}} & 90.04{\stdfont$\pm$.49} & 50.84{\stdfont$\pm$.93} & 69.36{\stdfont$\pm$.73} \\
    & Transfer & 63.93{\stdfont$\pm$.85} & 85.73{\stdfont$\pm$.60} & 50.62{\stdfont$\pm$.86} & 72.24{\stdfont$\pm$.65}\\
    & ReFine & 67.39{\stdfont$\pm$.89} & \textbf{90.96{\stdfont$\pm$.50}} & \textbf{51.21{\stdfont$\pm$.82}} & \textbf{74.39{\stdfont$\pm$.72}} \\
    \midrule
    & & \multicolumn{2}{c}{ISIC} & \multicolumn{2}{c}{ChestX} \\
    \midrule
    \multirow{3}{*}{\shortstack{miniImageNet}} & Linear    & 30.42{\stdfont$\pm$.54} & 42.97{\stdfont$\pm$.56} & 22.17{\stdfont$\pm$.37} & 25.80{\stdfont$\pm$.43} \\
    & Transfer & 32.31{\stdfont$\pm$.63} & 49.67{\stdfont$\pm$.62} & 21.82{\stdfont$\pm$.40} & 26.10{\stdfont$\pm$.44} \\
    & ReFine & \textbf{35.30{\stdfont$\pm$.59}} & \textbf{51.68{\stdfont$\pm$.63}} & \textbf{22.48{\stdfont$\pm$.41}} & \textbf{26.76{\stdfont$\pm$.42}} \\
    \midrule
    \multirow{3}{*}{\shortstack{tieredImageNet}} & Linear & 28.14{\stdfont$\pm$.55} & 37.20{\stdfont$\pm$.53} & 22.33{\stdfont$\pm$.40} & 25.03{\stdfont$\pm$.41} \\
    & Transfer & \textbf{32.31{\stdfont$\pm$.60}} & \textbf{46.36{\stdfont$\pm$.65}} & \textbf{22.49{\stdfont$\pm$.41}} & \textbf{25.76{\stdfont$\pm$.41}}\\
    & ReFine & 28.24{\stdfont$\pm$.48} & 38.83{\stdfont$\pm$.54} & 21.68{\stdfont$\pm$.36} & 24.83{\stdfont$\pm$.37} \\
    \bottomrule
    \end{tabular} 
}
\end{table}

\subsection{Experimental Setup}\label{subsec:setup}

\paragraph{Datasets.}
For the source domain dataset, we use miniImageNet (miniIN) \cite{matchingnet} and tieredImageNet (tieredIN) \cite{tieredimagenet}.
For the target domain, we use the BSCD-FSL benchmark \cite{bscd_fsl}, which consists of four different datasets: CropDisease \cite{cropdisease}, EuroSAT \cite{eurosat}, ISIC \cite{isic}, and ChestX \cite{chestx}, in order of similarity to miniIN.


\paragraph{Backbone and Training Setup.}
We use ResNet10 for miniIN and ResNet18 for tieredIN. Figure \ref{fig:backbone} describes the ResNet10 backbone. A family of ResNet consists of one stem module and four stages. The stem module consists of Conv-BN-ReLU-MaxPool layers.
Each stage includes one or more convolution blocks, where resolution is halved and the number of channels is doubled in the first block, and they are maintained in the following blocks.
For the pre-training and fine-tuning setups, we follow \citet{bscd_fsl} 

\begin{figure}[t]
     \centering
     \begin{subfigure}[h]{0.48\linewidth}
         \centering
         \includegraphics[width=\linewidth]{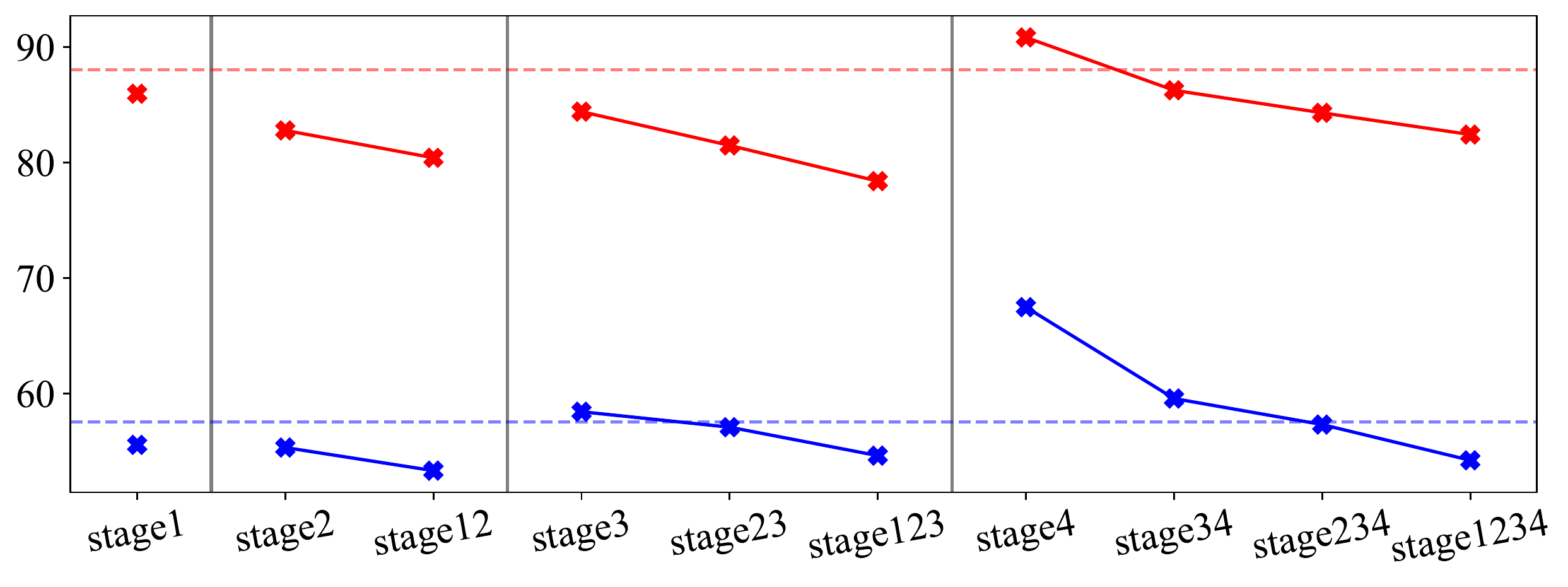}
         \caption{CropDisease}
         \label{fig:stagewise_CropDisease}
     \end{subfigure}
     \hfill
     \begin{subfigure}[h]{0.48\linewidth}
         \centering
         \includegraphics[width=\linewidth]{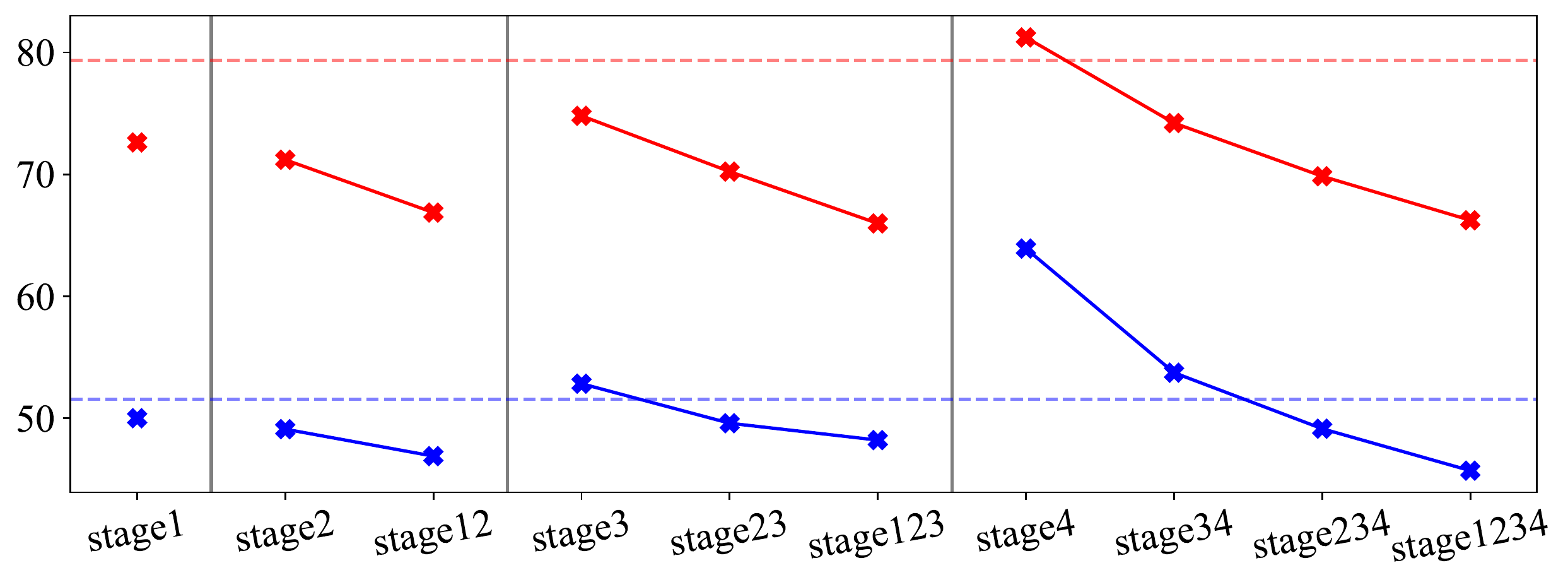}
         \caption{EuroSAT}
         \label{fig:stagewise_EuroSAT}
     \end{subfigure}
     
     \begin{subfigure}[h]{0.48\linewidth}
         \centering
         \includegraphics[width=\linewidth]{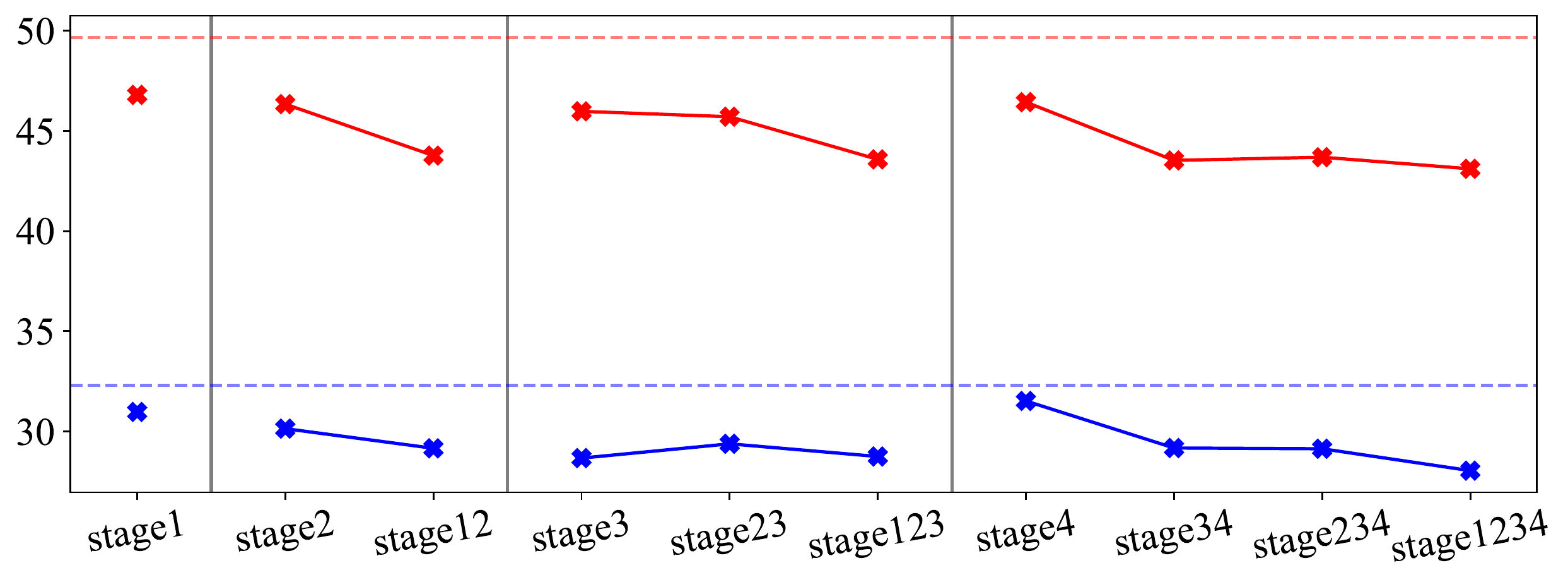}
         \caption{ISIC}
         \label{fig:stagewise_ISIC}
     \end{subfigure}
     \hfill
     \begin{subfigure}[h]{0.48\linewidth}
         \centering
         \includegraphics[width=\linewidth]{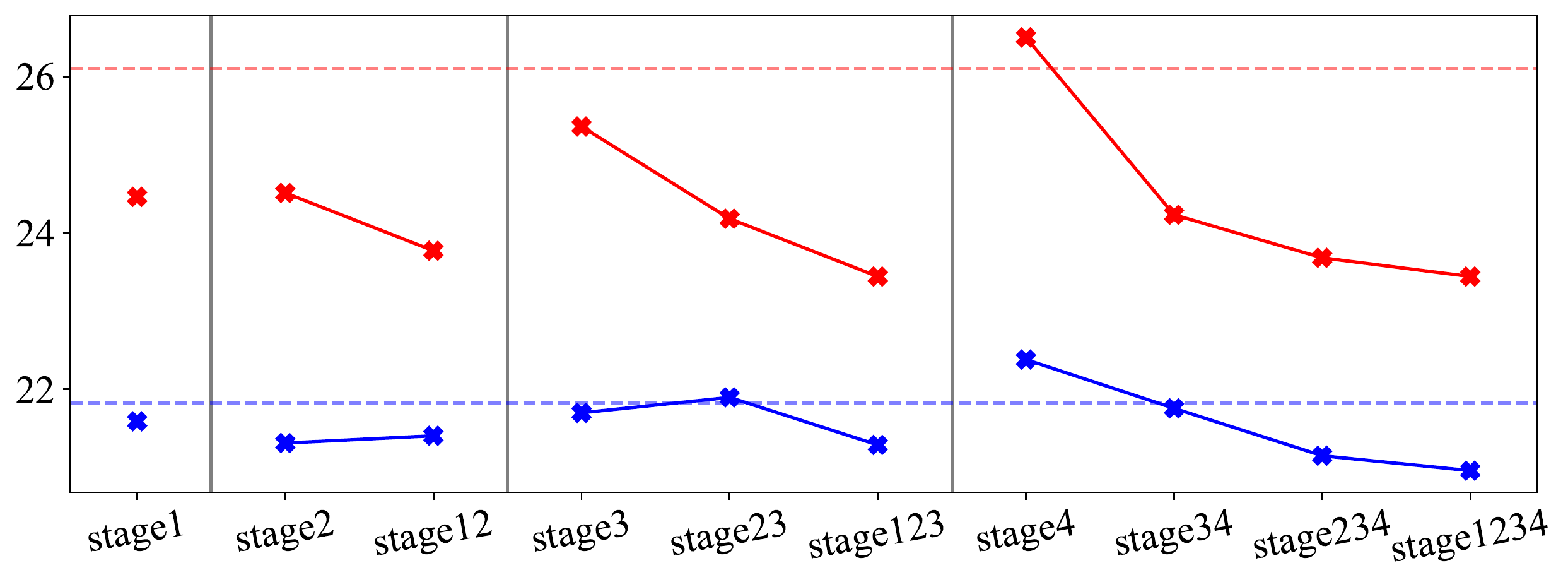}
         \caption{ChestX}
         \label{fig:stagewise_ChestX}
     \end{subfigure}
     \vspace{-5pt} 
     \caption{Accuracy trends according to the re-randomized stage(s). $x$-axis indicates re-randomized stage(s), and the blue and red lines indicate 1-/5-shot performance (\%), respectively. The dashed lines are the performances of Transfer.}
     \vspace{-10pt}
     \label{fig:stagewise}
\end{figure}

\begin{table}[t]
\caption{5-way $k$-shot accuracy over 600 tasks on \{miniIN\} $\rightarrow$ \{BSCD-FSL\} according to the parts of re-randomization in the last stage. The topmost layers are boldfaced.}
\centering
\addtolength{\tabcolsep}{-4pt}
\resizebox{\linewidth}{!}{%
\begin{tabular}{cc|ccccc|ccc|ccc}
    \toprule
    Path & Layer & \multicolumn{11}{c}{Re-randomization layer} \\
    \midrule 
    \multirow{4}{*}{Original}& Conv1        & \checkmark & & \checkmark & & \checkmark & & & & \checkmark & & \checkmark \\
     & BN1          & \checkmark & & & \checkmark & \checkmark & & & & \checkmark & & \checkmark \\
     & \textbf{Conv2}       & & \checkmark & \checkmark & & \checkmark & & & & & \checkmark & \checkmark \\
    & \textbf{BN2}         & & \checkmark & & \checkmark & \checkmark & & & & & \checkmark & \checkmark \\
    \midrule
    \multirow{2}{*}{ShortCut} & ShortCutConv& & & & & & \checkmark & & \checkmark & \checkmark & \checkmark & \checkmark \\
     & ShortCutBN  & & & & & & & \checkmark & \checkmark & \checkmark & \checkmark & \checkmark \\
    \midrule
    \multicolumn{13}{l}{1-shot} \\
    \midrule
	\multicolumn{2}{c|}{CropDisease}   & 66.85 & \textbf{68.93} & 62.34 & 62.43 & 67.74 & 52.60 & 64.32 & 58.22 & 66.41 & {68.83} & 67.74 \\
	\multicolumn{2}{c|}{EuroSAT}       & 61.99 & \textbf{64.14} & 59.19 & 58.91 & 62.96 & 53.03 & 56.29 & 57.33 & 62.01 & {62.77} & 63.90 \\
	\multicolumn{2}{c|}{ISIC}          & 31.91 & \textbf{35.30} & {34.32} & 30.96 & 32.88 & 29.42 & 32.45 & 30.22 & 30.55 & 33.21 & 31.17 \\
	\multicolumn{2}{c|}{ChestX}        & 22.20 & {22.48} & 22.00 & 21.53 & 22.08 & 21.19 & 21.55 & 21.60 & 21.99 & \textbf{22.82} & 22.46 \\
    \midrule
    \multicolumn{13}{l}{5-shot} \\
    \midrule
    \multicolumn{2}{c|}{CropDisease}   & 89.78 & 90.75 & 87.29 & 89.12 & 89.99 & 87.78 & 90.03 & 89.44 & 90.03 & \textbf{90.89} & {90.82} \\
    \multicolumn{2}{c|}{EuroSAT}       & 81.02 & \textbf{82.36} & 79.16 & 80.10 & 81.01 & 78.87 & 80.77 & 81.13 & 81.56 & {82.24} & 81.22 \\
    \multicolumn{2}{c|}{ISIC}          & 49.73 & {51.68} & \textbf{51.90} & 46.17 & 50.20 & 46.49 & 48.00 & 46.21 & 46.85 & 49.59 & 46.44 \\
    \multicolumn{2}{c|}{ChestX}        & 26.30 & \textbf{26.76} & 25.41 & 25.51 & 26.00 & 25.72 & 26.39 & 26.29 & 26.07 & {26.60} & 26.50 \\
    \bottomrule
    \end{tabular} 
}\label{tab:where_to_reinit_in_block}
\end{table}

\subsection{Performance Comparison}\label{subsubsec:only_rerand_comp}


Table \ref{tab:BSCD_comparison_all} describes the 5-way $k$-shot performance of Linear, Transfer, and ReFine in which a model is pre-trained on miniIN or tieredIN and then fine-tuned on BSCD-FSL. In most cases, ReFine outperforms Linear and Transfer. This implies that random parameters are generally better than the source-fitted parameters, especially of the topmost layers, for fine-tuning initialization.
Meanwhile, in the ISIC and ChestX data, we observed that it might be advantageous to transfer the source information to the target without re-randomization when the source data becomes larger.

\subsection{Ablation Study on Where to Re-randomize}\label{subsubsec:where_to_reinit}
We demonstrate that re-randomizing the extractor at the topmost layers is essential. We only consider Transfer as a baseline for fair comparison because ReFine fine-tunes the entire network.
Figure \ref{fig:stagewise} shows the test accuracy according to the re-randomized stage(s).

In Figure \ref{fig:stagewise}, we observe that re-randomizing only the last stage is the best. This is indicated by the performance decrease from re-randomizing more stages within each subdivision separated by vertical lines, and by the best performance in the rightmost subdivision when only one stage is re-randomized.


Furthermore, we investigate layer-wise re-randomization within the last stage for more granular analysis on where to re-randomize.
Table \ref{tab:where_to_reinit_in_block} describes the results according to the re-randomized layers in the last stage.
Re-randomizing only \{Conv2, BN2\} shows the best performance overall. We conclude that \emph{re-randomizing the topmost layers, excluding the shortcut path, in the last stage is a good rule of thumb}. A similar trend appears when the model is pre-trained on tieredIN, as described in Table \ref{tab:where_to_reinit_in_block_tiered}.

\begin{table}[!t]
\centering
\small
\caption{5-way $k$-shot accuracy over 600 tasks on \{tieredIN\} $\rightarrow$ \{BSCD-FSL\} according to the parts of re-randomization in the last stage. The topmost layers are boldfaced.}\label{tab:where_to_reinit_in_block_tiered}
\addtolength{\tabcolsep}{-4pt}{%
\resizebox{\linewidth}{!}{%
\begin{tabular}{c|cccccc|cccc}
    \toprule
    {} & \multicolumn{10}{c}{Re-randomization layer} \\
    \midrule 
    Block1.Conv1          & & & & & & \checkmark & & & & \checkmark \\
    Block1.BN1            & & & & & & \checkmark & & & & \checkmark \\
    Block1.Conv2          & & & & & \checkmark & \checkmark & & & \checkmark & \checkmark \\
    Block1.BN2            & & & & & \checkmark & \checkmark & & & \checkmark & \checkmark \\
    Block1.Conv3          & & & & \checkmark & \checkmark & \checkmark & & \checkmark & \checkmark & \checkmark \\
    Block1.BN3            & & & & \checkmark & \checkmark & \checkmark & & \checkmark & \checkmark & \checkmark \\
    Block1.ShortCutConv   & & & & & & & \checkmark & \checkmark & \checkmark & \checkmark \\
    Block1.ShortCutBN     & & & & & & & \checkmark & \checkmark & \checkmark & \checkmark \\
    Block2.Conv1          & & & \checkmark & \checkmark & \checkmark & \checkmark & \checkmark & \checkmark & \checkmark & \checkmark \\
    Block2.BN1            & & & \checkmark & \checkmark & \checkmark & \checkmark & \checkmark & \checkmark & \checkmark & \checkmark \\
    Block2.Conv2          & & \checkmark & \checkmark & \checkmark & \checkmark & \checkmark & \checkmark & \checkmark & \checkmark & \checkmark \\
    Block2.BN2            & & \checkmark & \checkmark & \checkmark & \checkmark & \checkmark & \checkmark & \checkmark & \checkmark & \checkmark \\
    \textbf{Block2.Conv3} & \checkmark & \checkmark & \checkmark & \checkmark & \checkmark & \checkmark & \checkmark & \checkmark & \checkmark & \checkmark \\
    \textbf{Block2.BN3}   & \checkmark & \checkmark & \checkmark & \checkmark & \checkmark & \checkmark & \checkmark & \checkmark & \checkmark & \checkmark \\
    \midrule
    \multicolumn{11}{l}{1-shot} \\
    \midrule
    CropDisease   & 67.39 & \textbf{68.31} & 60.98 & 52.84 & 48.43 & 42.61 & 51.82 & 52.78 & 51.37 & 49.28 \\
    EuroSAT       & \textbf{51.21} & 48.18 & 36.16 & 35.19 & 34.22 & 35.60 & 38.01 & 40.35 & 40.60 & 40.37 \\
    ISIC          & \textbf{28.24} & 28.06 & 27.02 & 26.64 & 26.12 & 26.94 & 26.24 & 26.35 & 26.42 & 26.70 \\
    ChestX        & \textbf{21.68} & 21.24 & 21.31 & 21.12 & 21.19 & 21.14 & 21.32 & 21.08 & 21.21 & 21.06 \\
    \midrule
    \multicolumn{11}{l}{5-shot} \\
    \midrule
	CropDisease   & \textbf{90.96} & 90.84 & 90.25 & 87.25 & 86.44 & 84.06 & 83.22 & 83.00 & 84.36 & 83.17 \\
	EuroSAT       & \textbf{74.39} & 74.03 & 71.54 & 67.58 & 66.26 & 62.66 & 60.22 & 62.07 & 63.17 & 60.40 \\
	ISIC          & 38.83 & 38.76 & 37.29 & 37.85 & 38.75 & 39.85 & 37.29 & 38.35 & 39.63 & \textbf{40.91} \\
	ChestX        & 24.83 & \textbf{24.90} & 24.64 & 24.08 & 23.66 & 23.54 & 23.23 & 22.88 & 23.15 & 22.89 \\
    \bottomrule
    \end{tabular}}
}
\end{table}

\subsection{Ablation Study on How to Re-randomize}\label{subsubsec:how_to_reinit}

\begin{table}[!t]
\caption{Analysis on the initializing distribution of ReFine. Sparse distribution initializes parameters with 20\% sparsity. Lottery indicates re-initialization.}
\label{tab:reinit_dist}
\resizebox{\linewidth}{!}{%
\begin{tabular}{c|l|cccc}
    \toprule
    Shot & Distribution & CropDisease & EuroSAT & ISIC & ChestX \\
    \midrule
    \multirow{5}{*}{1} & \textbf{Uniform} & 68.93{\stdfont $\pm$.84}& 64.14{\stdfont $\pm$.82} & 35.30{\stdfont $\pm$.59} & 22.48{\stdfont $\pm$.41} \\
    & Normal & 69.34{\stdfont $\pm$.86} & 60.85{\stdfont $\pm$.82} & 31.35{\stdfont $\pm$.58} & 22.38{\stdfont $\pm$.39} \\
    & Orthogonal & 67.96{\stdfont $\pm$.84} & 59.71{\stdfont $\pm$.83} & 31.05{\stdfont $\pm$.59} & 22.50{\stdfont $\pm$.38} \\
    & Sparse & 69.07{\stdfont $\pm$.84} & 61.21{\stdfont $\pm$.82} & 31.10{\stdfont $\pm$.61} & 22.52{\stdfont $\pm$.39} \\
    & Lottery & 61.53{\stdfont $\pm$.92} & 61.30{\stdfont $\pm$.88} & 31.27{\stdfont $\pm$.57} & 21.87{\stdfont $\pm$.36} \\
    \midrule
    \multirow{5}{*}{5} & \textbf{Uniform} & 90.75{\stdfont $\pm$.49} & 82.36{\stdfont $\pm$.57} & 51.68{\stdfont $\pm$.63} & 26.76{\stdfont $\pm$.42} \\
    & Normal & 91.31{\stdfont $\pm$.48} & 81.97{\stdfont $\pm$.58} & 46.92{\stdfont $\pm$.61} & 26.27{\stdfont $\pm$.43} \\
    & Orthogonal & 91.01{\stdfont $\pm$.50} & 81.92{\stdfont $\pm$.58} & 45.73{\stdfont $\pm$.59} & 26.71{\stdfont $\pm$.43} \\
    & Sparse & 91.33{\stdfont $\pm$.47} & 81.33{\stdfont $\pm$.60} & 45.62{\stdfont $\pm$.58} & 26.36{\stdfont $\pm$.43} \\
    & Lottery & 89.81{\stdfont $\pm$.50} & 81.96{\stdfont $\pm$.58} & 48.10{\stdfont $\pm$.62} & 26.45{\stdfont $\pm$.43} \\
    \bottomrule
\end{tabular}}
\end{table}


Table \ref{tab:reinit_dist} shows that re-randomizing the parameters following uniform distribution is generally the best practice. Uniform and Normal indicate that the values are sampled from the uniform and normal distribution. Orthogonal indicates the weights are randomized as an orthogonal matrix, as described in \citet{saxe2013exact}. Sparse indicates the weights are randomized as a sparse matrix, where non-zero elements are sampled from the zero-mean normal distribution, as described in \citet{martens2010deep}.
Lottery refers to re-initialization, i.e., resetting parameters to their initial state, prior to training. In the model pruning literature, the lottery ticket hypothesis \citep{frankle2018lottery} suggests that re-initialization can improve performance. However, we find that re-randomization is better suited in the case of CD-FSL. We believe that although re-initialization can be helpful in the original domain, this is not true under domain differences.





\section{Conclusion}\label{sec:conclusion}
We propose \textbf{ReFine} (\underline{Re}-randomization before \underline{Fine}-tuning), a simple yet effective method for CD-FSL, that involves resetting the parameters fitted to the source domain in order to maximize the efficacy of few-shot adaptation to the labeled target dataset.
We demonstrate that our method outperforms conventional baselines under the CD-FSL setup. Furthermore, we investigate where and how to re-randomize the pre-trained models.
We believe that our research will inspire CD-FSL researchers with the concept of removing information that is specific to the source domain.


\begin{acks}
This work was supported by Institute of Information \& communications Technology Planning \& Evaluation (IITP) grant funded by the Korea government(MSIT) (No.2022-0-00641, XVoice: Multi-Modal Voice Meta Learning; No.2019-0-00075, Development of AI Autonomy and Knowledge Enhancement for AI Agent Collaboration).
\end{acks}

\bibliographystyle{ACM-Reference-Format}
\balance
\bibliography{acmart}

\end{document}